%% file: sample-sigconf-authordraft.tex
\documentclass[sigconf, screen, table]{acmart}
\usepackage{pifont}
\AtBeginDocument{%
  }

\setcopyright{acmlicensed}
\copyrightyear{2025}
\acmYear{2025}
\acmDOI{XXXXXXX.XXXXXXX}
\acmConference[Conference acronym '25]{Make sure to enter the correct
  conference title from your rights confirmation email}{October 27–31,
  2025}{Dublin, Ireland}
\acmISBN{978-1-4503-XXXX-X/2018/06}

\renewcommand\footnotetextcopyrightpermission[1]{}

\acmSubmissionID{48}

\usepackage{hyperref} 

\usepackage{pifont}
\usepackage{graphicx}
\usepackage{wrapfig}
\usepackage{graphicx}
\usepackage{booktabs}
\usepackage{svg} % 需使用包
\usepackage{colortbl}  %彩色表格需要加载的宏包

\usepackage{multirow}
\usepackage{indentfirst}
\usepackage{threeparttable}
\usepackage{pifont}
\usepackage{makecell}
\usepackage{arydshln}
\usepackage{arydshln}
\usepackage{tabularx}
\newcommand{\blue}[1]{\textbf{\textcolor{mblue}{#1}}}
\newcommand{\red}[1]{\textbf{\textcolor{mred}{#1}}}
\usepackage{makecell}

\usepackage{pifont}
\usepackage{minitoc}
\definecolor{mblue}{RGB}{0, 77, 128}

\definecolor{mblue}{RGB}{0, 77, 128}
\definecolor{mred}{RGB}{192,0, 0}
\definecolor{darkgreen}{rgb}{0.0, 0.5, 0.0}
\definecolor{mycolor_blue}{HTML}{E7EFFA}

\definecolor{mycolor_gray}{HTML}{ECECEC}
\newcommand{\cmark}{\textcolor{darkgreen}{\ding{51}}}  % 深绿色的勾
\newcommand{\xmark}{\textcolor{red}{\ding{55}}}        % 红色的叉
\hypersetup{
    colorlinks=true,        % 如果想要彩色链接
    linkcolor=blue,        % 用于普通链接的颜色
    filecolor=magenta,     % 用于文件链接的颜色
    urlcolor=magenta,          % 用于URL的颜色
}

\begin{document}

%%
%% The "title" command has an optional parameter,
%% allowing the author to define a "short title" to be used in page headers.
\title{DFBench: Benchmarking Deepfake Image Detection Capability of Large Multimodal Models}
\author{
Jiarui Wang$^{1}$,    
Huiyu Duan$^{1}$, 
Juntong Wang$^{1}$,
Ziheng Jia$^{1}$,
Woo Yi Yang$^{1}$,
Xiaorong Zhu$^{1}$,
Yu Zhao$^{1}$,
Jiaying Qian$^{1}$,
Yuke Xing$^{1}$,
Guangtao Zhai$^{1}$,
Xiongkuo Min$^{1*}$
\\
% $^{1}$Institute of Image Communication and Network Engineering, \\
% $^{2}$ MoE Key Lab of Artificial Intelligence, AI Institute,\\ 
$^{1}$Shanghai Jiao Tong University, Shanghai, China. \\
% $^{2}$ Shanghai AI Lab.
}

\renewcommand{\shortauthors}{Jiarui Wang, et al.}

%%
%% The abstract is a short summary of the work to be presented in the
%% article.
\input{section/0_abstract}

%%
%% The code below is generated by the tool at http://dl.acm.org/ccs.cfm.
%% Please copy and paste the code instead of the example below.
%%


%%
%% Keywords. The author(s) should pick words that accurately describe
%% the work being presented. Separate the keywords with commas.

%% A "teaser" image appears between the author and affiliation
%% information and the body of the document, and typically spans the
%% page.
\begin{teaserfigure}
  \input{figures/overview}
\end{teaserfigure}

% \received{20 February 2007}
% \received[revised]{12 March 2009}
% \received[accepted]{5 June 2009}
\definecolor{mycolor_green}{HTML}{E6F8E0}
%%
%% This command processes the author and affiliation and title
%% information and builds the first part of the formatted document.
\maketitle

\input{section/1_Intro}
\input{section/2_related_work}
\input{section/3_database}
\input{section/4_Bench}
\input{section/6_conclusion}

%%
%% The acknowledgments section is defined using the "acks" environment
%% (and NOT an unnumbered section). This ensures the proper
%% identification of the section in the article metadata, and the
%% consistent spelling of the heading.
% \begin{acks}
% To Robert, for the bagels and explaining CMYK and color spaces.
% \end{acks}

%%
%% The next two lines define the bibliography style to be used, and
%% the bibliography file.
\bibliographystyle{ACM-Reference-Format}
\bibliography{main}

% \newpage
% \appendix
% \twocolumn[{\section*{\centering LMME3DHF: Benchmarking and Evaluating Multimodal 3D \\ Human Face Generation with LMMs (Supplemental Materials)\\}}\vspace*{10mm}]
% \input{appendix/0_overview}
% \input{appendix/1_generation_model}
% \input{appendix/2_subjective_experiment}
% \input{appendix/3_database}
% \input{appendix/4_train_loss}
% \input{appendix/5_implementions}

% \input{appendix/figures_tables/sal_map_example}
% \input{appendix/figures_tables/saliency_predict_example}

\end{document}

%% file: section/0_abstract.tex
\begin{abstract}
With the rapid advancement of generative models, the realism of AI-generated images has significantly improved, posing critical challenges for verifying digital content authenticity. Current deepfake detection methods often depend on datasets with limited generation models and content diversity that fail to keep pace with the evolving complexity and increasing realism of the AI-generated content. Large multimodal models (LMMs), widely adopted in various vision tasks, have demonstrated strong zero-shot capabilities, yet their potential in deepfake detection remains largely unexplored.
To bridge this gap, we present \textbf{DFBench}, a large-scale \underline{\textbf{D}}eep\underline{\textbf{F}}ake \textbf{Bench}mark featuring \textbf{(i) broad diversity}, including 540,000 images across real, AI-edited, and AI-generated content,  \textbf{(ii) latest model}, the fake images are generated by 12 state-of-the-art generation models, and \textbf{(iii)
bidirectional benchmarking and evaluating} for both the detection accuracy of deepfake detectors and the evasion capability of generative models. 
Based on DFBench, we propose \textbf{MoA-DF}, \underline{\textbf{M}}ixture \underline{\textbf{o}}f \underline{\textbf{A}}gents for \underline{\textbf{D}}eep\underline{\textbf{F}}ake detection, leveraging a combined probability strategy from multiple LMMs. 
% Comprehensive experiments demonstrate that state-of-the-art generative models produce images nearly indistinguishable from real ones, posing significant challenges to current detection systems.  
MoA-DF achieves state-of-the-art performance, further proving the effectiveness of leveraging LMMs for deepfake detection.
% DFBench thus serves as a valuable resource for advancing deepfake detection research and  highlights the pressing need for more generalizable, and interpretable detection approaches to counter increasingly sophisticated generative techniques. 
Database and codes are publicly available at \url{https://github.com/IntMeGroup/DFBench}.

% The growing realism of AI-generated images poses significant challenges to digital content authenticity, particularly on social media platforms where deepfakes can spread rapidly and deceive large audiences. Existing deepfake detection efforts often rely on limited datasets and one-sided evaluation protocols that fail to capture the evolving complexity of generative models. To address these gaps, we introduce DFBench, a large-scale benchmark consisting of 540,000 images spanning real, synthetic, and tampered content. The fake images are produced using 12 state-of-the-art generation models, enabling a comprehensive and diverse evaluation landscape. Unlike prior datasets, DFBench adopts a dual evaluation paradigm: it assesses both the detection accuracy of large multimodal models and traditional detectors, and the perceptual realism of generated images by measuring their ability to evade detection. Experimental results reveal that many advanced generation techniques can produce images that are visually indistinguishable from real ones and highly deceptive to existing detection models. DFBench provides a valuable resource for advancing deepfake detection research and highlights the need for more robust and interpretable detection frameworks in the face of increasingly sophisticated generative models.
\end{abstract}

%% file: figures/overview.tex
\centering
\includegraphics[width=1\linewidth]{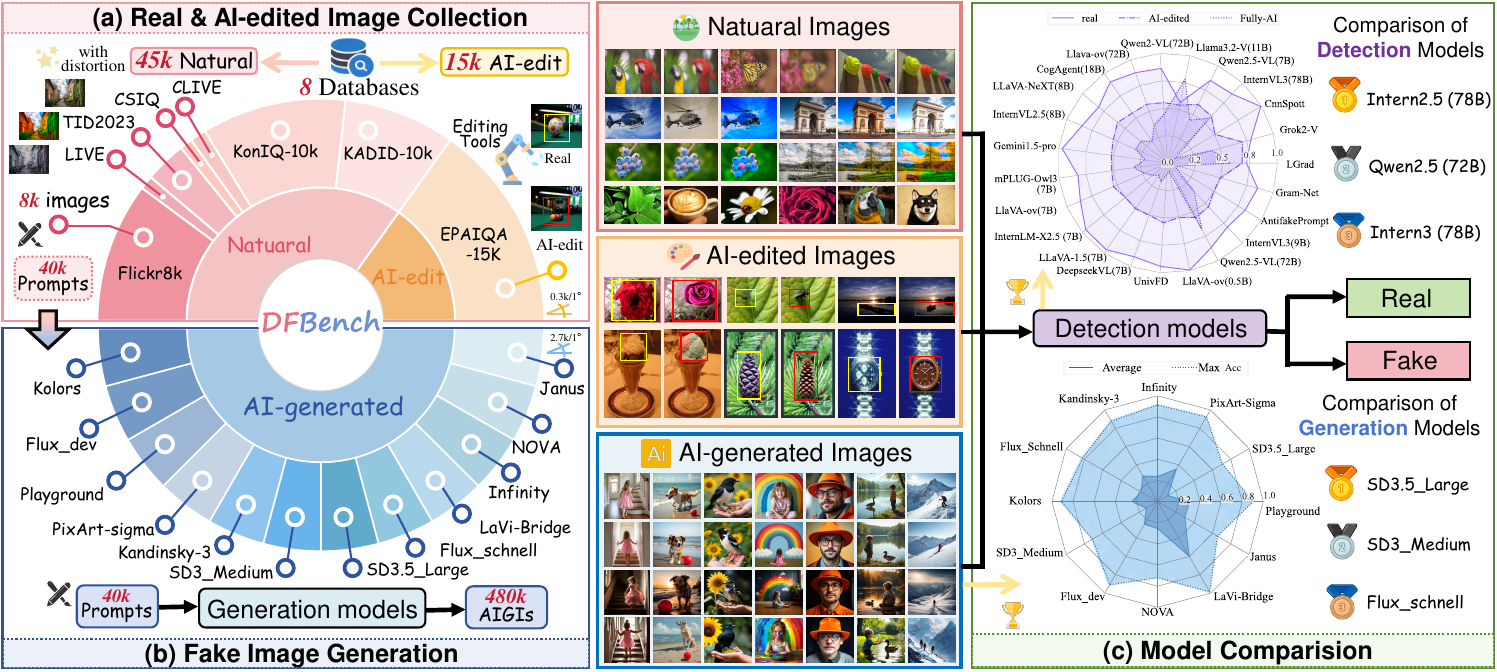}
\vspace{-8mm}
\caption{We present the DFBench, a large dataset for benchmarking deepfake image detection capabilities.
(a) 45K real and 15K AI-edited images are collected from 8 sources.
(b) 480K fake images are generated using 12 state-of-the-art generation models based on 40K prompts from Flickr8k.
(c) The database enables evaluation for both detection models and generation models.}
\label{overview}
\vspace{5mm}

%% file: section/1_intro.tex
\section{Introduction}
\input{tables/database}
The rapid advancement of generative models~\cite{li2024playground,chen2024pixart,kolors,flux2024,esser2024scaling,wu2024janus,zhao2024bridging,arkhipkin2024kandinsky} has significantly improved the ability to generate highly realistic images. However, these advancements raise serious concerns of the generated images regarding misinformation, social manipulation, and erosion of public trust. These concerns have driven the development of deepfake detection models~\cite{DBLP:journals/corr/abs-2310-17419,DBLP:journals/corr/abs-2104-02984,DBLP:conf/icml/FrankESFKH20,DBLP:conf/icip/JuJKXNL22,DBLP:conf/cvpr/LiuQT20,DBLP:conf/cvpr/OjhaLL23,DBLP:conf/cvpr/Tan0WGW23,DBLP:journals/corr/abs-2311-00962}. These models are typically trained on datasets containing real and fake images~\cite{yang2019exposing,DBLP:journals/corr/abs-2309-02218,gandhi2020adversarial,sha2022fake,bird2023cifake}, with the goal of distinguishing between real and fake content. Thus, the generalization ability of these deepfake image detection models remains questionable.
% Consequently, the quality and diversity of the datasets used for training and evaluation play a crucial role in achieving high-accuracy deepfake detection. 
% Well-designed datasets can expose models to subtle artifacts and diverse manipulation strategies, improving robustness and generalization in real-world scenarios.

Existing deepfake detection datasets and benchmarks~\cite{wang2019fakespotter,dang2020detection,DBLP:journals/corr/abs-2309-02218,gandhi2020adversarial,wang2023benchmarking} exhibit several critical limitations:
\textbf{(1) Limited generative models}: most datasets~\cite{yang2019exposing,DBLP:journals/corr/abs-2309-02218,gandhi2020adversarial,sha2022fake,bird2023cifake} rely on a small number of generative methods. Detection models trained on such datasets tend to learn model-specific artifacts, rather than generalizable features indicative of image manipulation. Moreover, many of the generative models~\cite{karras2017progressive,brock2018large,nichol2021glide,gu2022vector,dhariwal2021diffusion} used in earlier datasets are now outdated, often producing images with visible distortions, unnatural textures, or structural inconsistencies~\cite{wang2023aigciqa2023,wang2025lmm4lmmbenchmarkingevaluatinglargemultimodal,wang2025qualityassessmentaigenerated,wang2025tdveassessorbenchmarkingevaluatingquality,wang2024aigv,xu2025harmonyiqapioneeringbenchmarkmodel,wang2025lovebenchmarkingevaluatingtexttovideo,NEURIPS2024_46b5405a}, which makes detection relatively easy for modern detectors and hinder the generalization of detection models to newer and more complex fakes generated by state-of-the-art models~\cite{li2024playground,chen2024pixart,kolors,flux2024}.
\textbf{(2) Limited content diversity}: existing datasets for deepfake detection focus mainly on facial imagery~\cite{yang2019exposing,wang2019fakespotter,dang2020detection,DBLP:journals/corr/abs-2309-02218,gandhi2020adversarial,yang2025lmme3dhfbenchmarkingevaluatingmultimodal}, overlooking the growing threat of non-facial manipulations. In addition, most datasets include either fully real or fully fake images~\cite{verdoliva2022,wang2023benchmarking,sha2022fake}, lacking examples of partially AI-edited content where only specific regions are manipulated~\cite{qian2025explainablepartialaigcimagequality,Diffedit}. Furthermore, the real images in current datasets are often clean and undistorted~\cite{DBLP:journals/ijcv/PlummerWCCHL17,DBLP:conf/eccv/LinMBHPRDZ14}, making detection easier. In real scenarios, authentic images frequently contain compression artifacts, motion blur, or other imperfections~\cite{sheikh2006statistical,duan2024finevq,ponomarenko2015image,lin2019kadid}. Including these natural distortions is important to improve the robustness of detection benchmarks.
\textbf{(3) Limited evaluation scope}: current benchmarks mainly focus on evaluating the performance of dedicated deepfake detection models. 
Large multimodal models (LMMs) have demonstrated strong zero-shot capabilities in vision tasks, yet their potential in deepfake detection remains largely unexplored.
% without systematically benchmarking the detection resilience of large multimodal models (LMMs). 
% Moreover, as generative models become increasingly advanced, evaluating their success should not rely solely on visual quality, but also on their ability to deceive existing detection systems. Therefore, it is equally important to assess the \textit{\textbf{evasion capability}} of the generative models.
% A bidirectional benchmark is needed to evaluate both the detection ability of detectors and the evasion ability of generative models.

DFBench is specifically designed to overcome the key limitations of existing datasets.
(1) To improve generative diversity, fake images are generated using \textbf{12 state-of-the-art models}, covering a wide range of generation contents.
(2) DFBench enhances content diversity by including \textbf{partially manipulated images} where only specific regions are edited and real images with natural distortions (\textit{e.g.}, blur, compression) to better reflect real-world scenarios.
(3) DFBench adopts a \textbf{bidirectional evaluation protocol} that assesses both the detection ability of conventional detectors and LMMs, and the evasion ability of generative models in fooling these detectors.
% , providing a more comprehensive evaluation of both detection robustness and generation realism.
As shown in Figure~\ref{overview}, DFBench includes highly deceptive examples that challenge deepfake detection models. Table~\ref{tab:comparison} further highlights its advantages in scale, diversity, and evaluation design compared to existing benchmarks.
Based on DFBench, we propose \textbf{MoA-DF}, \underline{\textbf{M}}ixture \underline{\textbf{o}}f \underline{\textbf{A}}gents for \underline{\textbf{D}}eep\underline{\textbf{F}}ake detection, leveraging a combined probability strategy from multiple LMMs and achieves state-
of-the-art performance, proving the effectiveness of LMMs in deepfake detection tasks. 
In summary, our main contributions are:
\vspace{-1mm}
\begin{itemize}
\item We introduce \textbf{DFBench}, a large-scale and diverse benchmark, featuring the \textbf{largest scale} of fake images generated by 12 state-of-the-art generative models, and \textbf{rich content} including AI-edited images and real-world image distortions (\textit{e.g.}, blur, noise, compression, color distortions).
% \item We enrich content diversity by incorporating \textbf{partially manipulated images}, a variety of \textbf{real-world image distortions} (\textit{e.g.}, blur, compression artifacts) from 8 data sources, and \textbf{high-quality synthetic images} produced by different generation models.
\item We present a \textbf{bidirectional evaluation protocol} that benchmarks both the \textit{\textbf{detection accuracy}} of deepfake detectors and the \textit{\textbf{evasion capability}} of generative models.
% , enabling a comprehensive understanding of the interplay between generation and detection.
\item We propose \textbf{MoA-DF}, a novel mixture of agents method that combines the probabilistic outputs of LMMs to achieve more robust and accurate deepfake detection.
\end{itemize}
\vspace{-2mm}

%% file: tables/database.tex
\begin{table*}[t]
	\centering
	\small

     \vspace{-2mm}
    \renewcommand\arraystretch{0.85}
	\caption{An overview of fake image detection datasets.}
    \vspace{-4mm}
		\resizebox{\linewidth}{!}{\begin{tabular}{ccccccccc}
			\hline
			\multirow{2}{*}{\textbf{Dataset}}                                     & \multirow{2}{*}{\begin{tabular}[c]{@{}c@{}}\textbf{Image} \\ \textbf{Content}\end{tabular}} & \multicolumn{2}{c}{\textbf{AI Generation Category}}                                     & \multirow{2}{*}{\begin{tabular}[c]{@{}c@{}}\textbf{Public}\\ \textbf{Avalibility}\end{tabular}}  &\multirow{2}{*}{\begin{tabular}[c]{@{}c@{}}\textbf{Database}\\\textbf{Real Sources}\end{tabular}} & \multirow{2}{*}{\begin{tabular}[c]{@{}c@{}}\textbf{AI}  \\ \textbf{Models}\end{tabular}} & \multirow{2}{*}{\begin{tabular}[c]{@{}c@{}}\textbf{Fake }\\ \textbf{Images}\end{tabular}} & \multirow{2}{*}{\begin{tabular}[c]{@{}c@{}}\textbf{Total} \\ \textbf{Images}\end{tabular}}\\ \cline{3-4}
			&                                                                           & Fully AI Generation                     & Partial AI Editing                                      &                                                                               &                                                                         &                                                                         \\ \hline
			UADFV~\cite{yang2019exposing}          & Face                                                                      & \cmark & \xmark                     & \xmark                                                 &Real face & 1  & 252                                                                     & 493                                                                     \\
			FakeSpotter~\cite{wang2019fakespotter} & Face                                                                     & \cmark & \cmark                     & \xmark   &  CelebA, FFHQ & 8                                               & 5,000                                                                   & 11,000                                                                   \\
            
			DFFD~\cite{dang2020detection}          & Face                                                                      &\cmark & \xmark                     & \cmark                                    &CelebA, FFHQ, FaceForensics++ &        4        & 240,336                                                                  & 299,039                                                               \\
            DeepFakeFace~\cite{DBLP:journals/corr/abs-2309-02218}& Face &\cmark & \xmark&\cmark & IMDB-WIKI     & 3 &90,000 &120,000 \\
			APFDD~\cite{gandhi2020adversarial}     & Face                                                                      & \cmark & \xmark                     & \xmark          &     CelebA     &1                                 & 5,000                                                                   & 10,000                                                                   \\
			% ForgeryNet~\cite{he2021forgerynet}     & Face                                                                      & \cmark & \xmark                     & \cmark                                                    & 1,438,201                                                               & 1,457,861                                                               \\
			DeepArt~\cite{wang2023benchmarking}    & Art                                                                       &\cmark & \xmark                     & \cmark                 &LAION-5B &5                                   & 73,411                                                                  & 137,890                                                                 \\
			% CNNSpot~\cite{wang2020cnn}             & Object                                                                    &\cmark & \xmark                     & \cmark        &    SDXL &                      & 362,000                                                                 & 362,000                                                                 \\
			IEEE VIP Cup~\cite{verdoliva2022}      & General                                                                   &\cmark & \xmark                     & \xmark          &FFHQ,   Imagenet,         COCO,  LSUN & 5                             & 7,000                                                                   & 14,000                                                                   \\
			DE-FAKE~\cite{sha2022fake}             & General                                                                   & \cmark & \xmark                     & \xmark            &MSCOCO,     Flickr30k& 2                                    & 60,000                                                                  & 80,000                                                                  \\
			CiFAKE~\cite{bird2023cifake}           & General                                                                    & \cmark & \multicolumn{1}{c}{\xmark} & \cmark      & CIFAR      &   1                                     & 60,000                                                                  & 120,000    \\  
            SID-Set~\cite{huang2025sidasocialmediaimage}& General                                                                   & \cmark & \multicolumn{1}{c}{\cmark} & \cmark   &  COCO, Flickr30k, MagicBrush &1                                               & 200,000                                                                  &300,000   \\
			% \textbf{GenImage}                                                         & \textbf{General}                                                                   & \cmark & \multicolumn{1}{c}{\cmark} & \cmark                                                    & \textbf{1,331,167}                                                               & \textbf{1,350,000}     
            \rowcolor{gray!20} \textbf{DFBench (Ours)}  & General                                                                   & \cmark & \multicolumn{1}{c}{\cmark} & \cmark      & \textbf{8 Datasets }                      & \textbf{12 }                      & \textbf{495,000}                                                                  & \textbf{540,000}     
            \\ \hline
		\end{tabular}}
	
	\label{tab:comparison}
\vspace{-2mm}
\end{table*}

%% file: section/2_related_work.tex
\vspace{-1mm}
\section{RELATED WORK}
\input{figures/exam}
\input{figures/t2}
\subsection{Fake Image Generators }

\textbf{Generative Adversarial Networks (GANs)} such as WGAN~\cite{arjovsky2017wasserstein}, CycleGAN~\cite{zhu2017unpaired}, StyleGAN3~\cite{karras2021alias}, and GigaGAN~\cite{kang2023gigagan} are early models widely used for image synthesis.  However, GANs often suffer from training instability, mode collapse, and resolution limitations. 
\textbf{Diffusion models} recently emerge as a new paradigm for high-fidelity image generation~\cite{sohl2015deep,ho2020denoising}. Notable diffusion-based models such as ADM~\cite{dhariwal2021diffusion}, GLIDE~\cite{nichol2021glide}, and Stable Diffusion~\cite{Rombach_2022_CVPR} (SD) demonstrate superior image quality, semantic alignment, and robustness compared to GANs. The diffusion framework has been further extended to support high-resolution image synthesis by models like SDXL~\cite{podell2023sdxl}, PixArt-sigma~\cite{chen2024pixart}, Playground~\cite{li2024playground}, and Kolors~\cite{kolors}. These models also show improved controllability and are less prone to training collapse, making them suitable for constructing high-quality datasets.
\textbf{Autoregressive models (AR)} also gain traction in image generation due to their strong expressiveness and unification with large-scale language modeling. Janus~\cite{wu2024janus}, NOVA~\cite{deng2024nova}, Infinity~\cite{Infinity}, and EMU3~\cite{wang2024emu3} introduce efficient decoding strategies and improved spatial-temporal coherence. Although AR models typically suffer from slower inference due to sequential decoding, they offer flexible integration with language models.
% and strong generalization capabilities.

\vspace{-2mm}

\subsection{DeepFake Image Detection Datasets. }
\input{figures/model}
A variety of datasets have been developed to advance deepfake image detection. Early datasets like UADFV~\cite{yang2019exposing}, FakeSpotter~\cite{wang2019fakespotter}, and DFFD~\cite{dang2020detection} focused mainly on facial forgeries generated by classical GANs or simple editing, but are limited in both scale and diversity. Later works such as DeepFakeFace~\cite{DBLP:journals/corr/abs-2309-02218} and APFDD~\cite{gandhi2020adversarial} remained face-centric, while DeepArt~\cite{wang2023benchmarking} and DE-FAKE~\cite{sha2022fake} explored artistic or caption-driven generation. Datasets like CiFAKE~\cite{bird2023cifake} and IEEE VIP Cup~\cite{verdoliva2022} attempt broader coverage but often rely on low-resolution images or limited models.
Most existing datasets focus solely on fully generated images, with few considering partially edited cases. SID-Set~\cite{huang2025sidasocialmediaimage} introduces partial edits but uses only a single generation model and is still limited in scale. Additionally, the real images in most datasets are typically pristine and artifact-free, failing to represent naturally distorted content that detectors would encounter in the wild (\textit{e.g.}, compression artifacts, motion blur).
Our dataset DFBench stands out by its largest scale and broad diversity, including real, AI-edited, and fully generated images constructed from 8 sources.
% and generated from 12 latest models.

% \subsection{Saliency Prediction}
% Recent advances in computer vision have driven significant progress in human visual attention analysis, with saliency prediction emerging as a particularly crucial research area. This technology simulates the human visual system's mechanisms for allocating attention across different image regions. In recent years, numerous research efforts have focused on developing saliency prediction models. Traditional models \cite{bruce2005saliency, goferman2011context, erdem2013visual} primarily rely on low-level features such as intensity, color, and contrast to generate saliency maps. In contrast, deep learning-based models \cite{huang2015salicon, che2019gaze, lou2022transalnet} are capable of capturing more complex visual patterns and semantic information, allowing them to outperform traditional approaches across a range of tasks. However, despite their success in general applications, these models often fall short in the context of quality assessment, where identifying distortion-specific regions is crucial for accurately evaluating perceptual quality. This motivates us to develop a saliency prediction model specifically tailored for identifying distortion-aware salient regions in AI-generated human faces as distortion-aware saliency prediction not only guides model attention toward visually critical regions but also enhances interpretability by highlighting the exact areas that degrade user experience.

%% file: figures/exam.tex
\begin{figure*}[t]
    \centering
    \includegraphics[width=\linewidth]{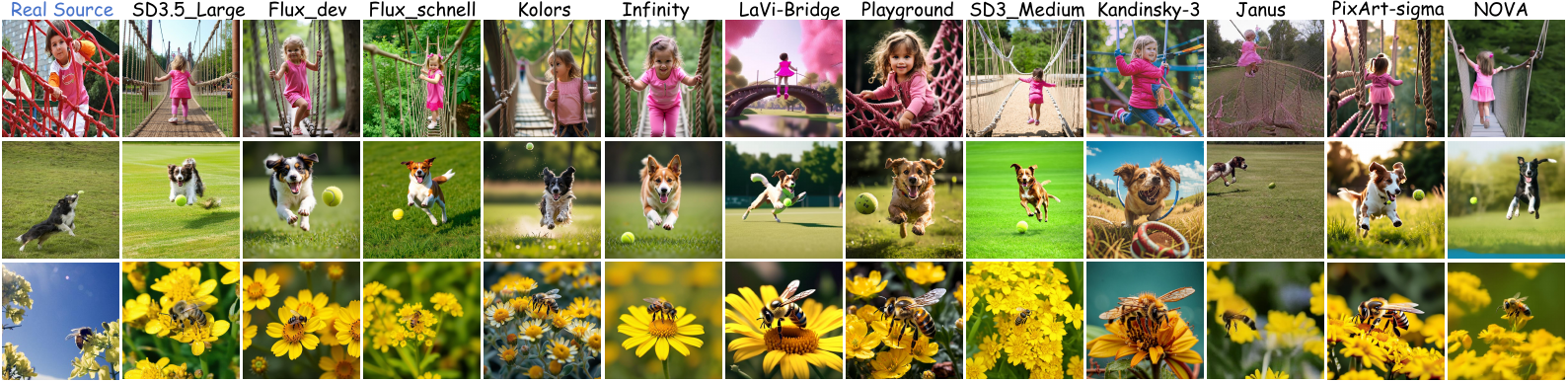}
     \vspace{-8mm}
    \caption{Visualization of images on the DFBench dataset.} 
     \vspace{-3mm}
    \label{example}
\end{figure*}

%% file: figures/t2.tex
\begin{figure*}[t]
    \centering
    \includegraphics[width=\linewidth]{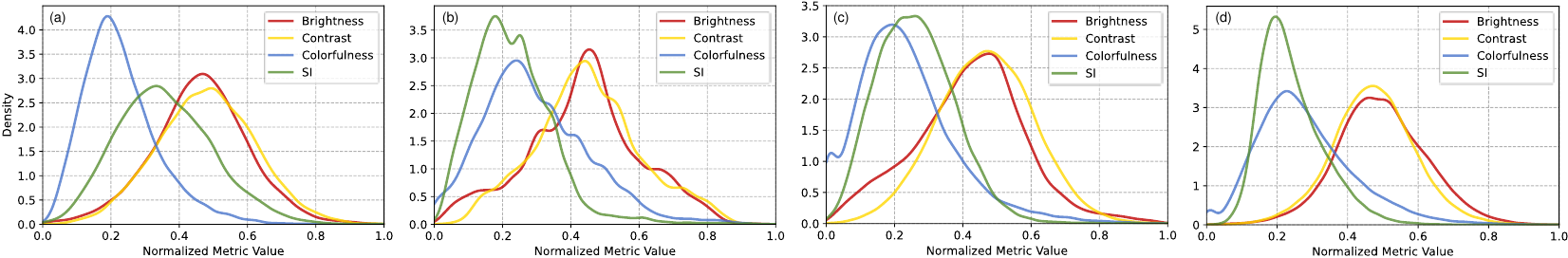}
     \vspace{-7mm}
    \caption{Feature distribution of the DFBench. (a) Feature distribution of real images with no distortion. (b) Feature distribution of real images with distortions. (c) Feature distribution of AI-edited  images. (d) Feature distribution of AI-generated images. } 
     \vspace{-2mm}
    \label{t222}
\end{figure*}

%% file: figures/model.tex
\begin{figure}
    \centering
    % \vspace{-5mm}
    \includegraphics[width=1\linewidth]{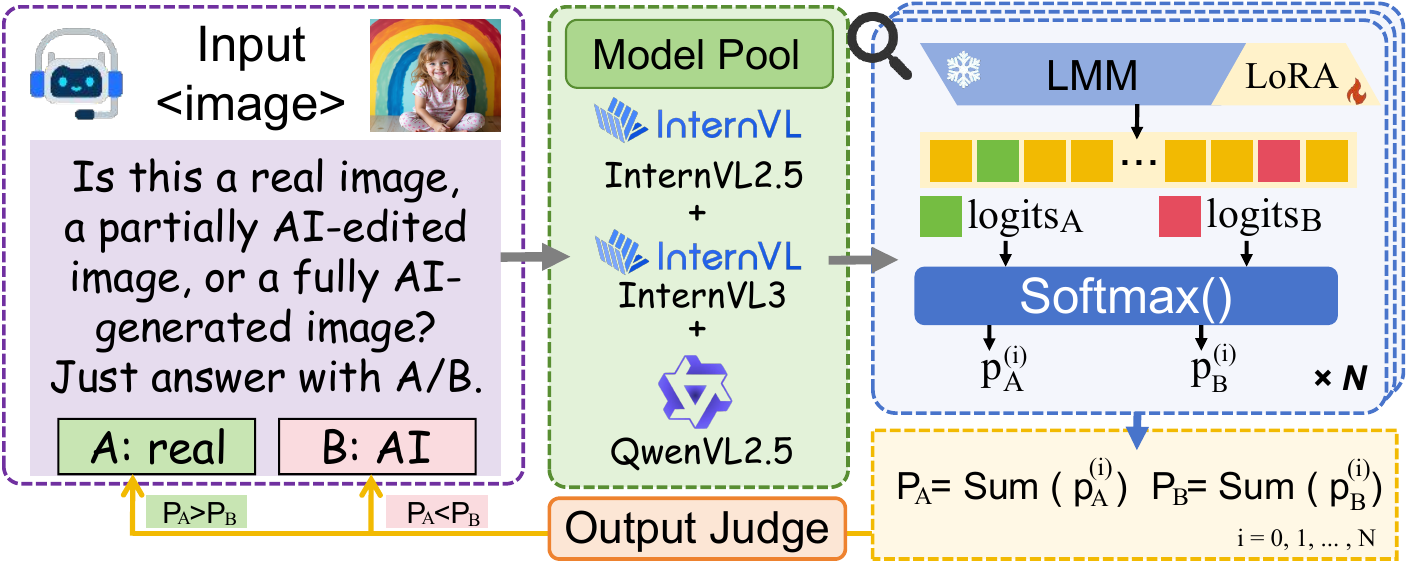}
    \vspace{-7mm}
    \caption{Overview of the MoA-DF architecture. Three LMMs are chosen as core detectors. Each model independently produces log-probabilities corresponding to the likelihood of the input image belonging to A (real) or B (fake). These log-probabilities are converted into normalized probabilities via the softmax function. The final decision is made based on the aggregation of these probabilities across all models. }
     \vspace{-9mm}
    \label{model}
\end{figure}

%% file: section/3_database.tex
% \input{figures/distortion}
\input{tables/real2}
\input{tables/partial}
\input{tables/Acc}
\vspace{-3mm}
\section{Database Construction}
\subsection{Image Collection and Generation}
% \subsection{Real and AI-edited Image Collection}
To ensure content diversity and realism, DFBench incorporates real images from seven well-known public natural image datasets, including  LIVE~\cite{sheikh2006statistical}, CSIQ~\cite{larson2010most}, TID2013~\cite{ponomarenko2015image}, KADID-10k~\cite{lin2019kadid}, CLIVE~\cite{ghadiyaram2015massive}, KonIQ-10k~\cite{hosu2020koniq}, and Flickr8k~\cite{hodosh2013framing}. 
% Among these, LIVE, CSIQ, TID2023, and KADID-10k primarily contain synthetically distorted images. 
% LIVE~\cite{sheikh2006statistical} contains 982 images with five distortion types. CSIQ~\cite{larson2010most} consists of 866 distorted images encompassing six distortion categories. TID2013~\cite{ponomarenko2015image} expands this diversity with 25 distinct distortion types.
% % ranging from noise, blur, color distortions to various compression artifacts. 
% KADID-10k~\cite{lin2019kadid} is a large-scale database with a total of 10,125 distorted images and 25 types of artificial distortions at multiple severity levels.
% % In contrast, CLIVE and KonIQ-10k consist of authentically distorted real-world images, reflecting naturally occurring degradations. 
% CLIVE~\cite{ghadiyaram2015massive} and KonIQ-10k~\cite{hosu2020koniq} comprise 1,162 and 10,073 real-world images, both focusing on authentic, naturally occurring distortions and diverse content scenarios. Flickr8k~\cite{hodosh2013framing} provides 8,000 undistorted images covering a broad array of general topics such as people, animals, and landscapes.
Except for Flickr8k, all other datasets include images affected by various forms of degradation that simulate real-world image quality impairments, including compression artifacts, blur, noise, color distortions, and \textit{etc.} 
% The inclusion of such distorted images is critical for challenging detection models, improving their robustness by compelling them to distinguish high-quality authentic images from subtly manipulated or degraded ones. 
The AI-edited images are from EPAIQA-15K~\cite{qian2025explainablepartialaigcimagequality}.
% , categorized into four editing types: style change (modifications to the visual style of the image), object enhancement (change appearance of specific objects), object operation (add, remove, or relocate objects), and semantic change (alterations that affect the image's context or meaning).  
% This approach enhances the detection model’s ability to perform reliably under diverse and realistic conditions.
% \input{tables/real2}
% \input{tables/partial}
% \input{tables/Acc}
% covering both general scenes and quality-degraded content. Specifically, we select high-quality reference images and authentic photographs from  Flickr8k~\cite{hodosh2013framing}. To introduce natural distortions, we include reference images and their degraded counterparts from standard IQA datasets such as LIVE~\cite{sheikh2006statistical}, CSIQ~\cite{larson2010most}, TID2023~\cite{ponomarenko2023tid2023}, KADID-10k~\cite{lin2019kadid}, CLIVE~\cite{ghadiyaram2015massive}, and KonIQ-10k~\cite{hosu2020koniq}. These real images represent various resolution levels, distortion types (e.g., blur, noise, compression), and semantic scenes, offering a robust basis for realistic deepfake construction.
% \vspace{-2mm}
% \subsection{Fake Image Generation}
To construct a diverse and challenging set of fake images, we utilize 12 state-of-the-art open-source image generation models, including 10 diffusion-based models: PixArt-sigma~\cite{chen2024pixart}, Playground~\cite{li2024playground}, Kolors~\cite{kolors}, SD3.5-Large~\cite{esser2024scaling}, SD3-Medium~\cite{esser2024scaling}, LaVi-Bridge~\cite{zhao2024bridging}, Kandinsky-3~\cite{arkhipkin2024kandinsky}, Flux-schnell~\cite{flux2024}, Flux-dev~\cite{flux2024}, Janus~\cite{wu2024janus}, and two AR-based models: NOVA~\cite{deng2024nova} and Infinity~\cite{Infinity}. To maintain fairness, all generative models are employed using their official default weights without further adaptation or tuning.
Using 40K prompts from Flickr8k~\cite{hodosh2013framing}, we generated a total of 480K images (12 models × 40,000 images).  Each of the 12 models is provided with the same set of prompts from real-world image captions, as shown in Figure~\ref{example}. Notably, models such as SD3.5-Large~\cite{esser2024scaling} and Flux-dev~\cite{flux2024} are capable of producing highly detailed outputs that even surpass the real source images,
% . These outputs can appear exceptionally realistic to human observers, thus 
posing substantial challenges to deepfake detection models. 
% This structured approach allows for the creation of a diverse and challenging dataset, which is critical for testing the robustness of deepfake detection models under various types of manipulation.

\vspace{-3mm}
\subsection{Database Analysis}

As illustrated in Figure~\ref{t222}, we analyze the feature distribution of real distortion-free, real distorted, AI-edited, and AI-generated images in the DFBench across four image quality-related features, including colorfulness, brightness, contrast, and spatial information (SI).
It can be observed that distorted real images generally exhibit lower colorfulness and higher SI values compared to distortion-free real images, likely due to the presence of noise, blur, or compression artifacts. AI-generated images exhibit the highest SI, reflecting their rich spatial detail.
AI-edited images exhibit feature values between real and synthetic content, due to their mixed authentic and manipulated content. The broad range of feature distributions establishes DFBench as a comprehensive benchmark for evaluating deepfake detection under realistic and challenging conditions.

\vspace{-2mm}
\section{The MoA-DF Method}
To leverage the strong zero-shot capabilities of LMMs for robust deepfake detection, we propose the MoA-DF, mixture of agents for deepfake detection that integrates the knowledge of multiple state-of-the-art LMMs. Specifically, we select Qwen2.5 (7B), InternVL2.5 (8B), and InternVL3 (9B) as the core detection agents. Each model outputs log-probabilities of A (real) or B (fake), denoted as \(\log p_i(A)\) and \(\log p_i(B)\) for model \(i\), which are then normalized using softmax: 
\begin{equation}
p_A^{(i)} = \frac{e^{\log p_A}}{e^{\log p_A} + e^{\log p_B},}, \quad
p_B^{(i)} = \frac{e^{\log p_B}}{e^{\log p_A} + e^{\log p_B}}
\end{equation}
% where \(p_A^{(i)}\) and \(p_B^{(i)}\) represent the probabilities predicted by the \(i\)-th model.
We then aggregate the predictions from all \(N=3\) models:
% We select Qwen2.5, InternVL2.5, and InternVL3 as the core detectors to form our ensemble framework. Each model independently outputs the log-probabilities of classifying an input image as either class A (real) or class B (fake). Specifically, for each model, we extract the log-probabilities \(\log p_A\) and \(\log p_B\) corresponding to classes A and B from the model output.
%  The probabilities are then obtained by applying a softmax function over the log-probabilities:
% \begin{equation}
% \mathbf{p} = \mathrm{softmax}(\mathbf{l}) = \left[\frac{e^{\log p_A}}{e^{\log p_A} + e^{\log p_B}}, \quad \frac{e^{\log p_B}}{e^{\log p_A} + e^{\log p_B}}\right]
% \end{equation}
% % where \(p_A\) and \(p_B\) denote the predicted probabilities for classes A and B, respectively.
% After obtaining \((p_A, p_B)\) from each of the three models, we aggregate their outputs by summing the probabilities across models to form the ensemble score:
\vspace{-2mm}
\begin{equation}
P_A = \sum_{i=1}^{N} p_A^{(i)}, \quad P_B = \sum_{i=1}^{N} p_B^{(i)}
\end{equation}
The final decision \(D\) is made:
\begin{equation}
D = 
\begin{cases}
\text{A (Real)}, & \text{if } P_A > P_B \\
\text{B (Fake)}, & \text{otherwise}
\end{cases}
\end{equation}
This ensemble strategy effectively leverages the diverse strengths and perspectives of multiple large models by fusing their soft predictions. By combining probabilistic outputs, MoA-DF mitigates individual model biases and uncertainties, resulting in enhanced robustness and improved overall detection accuracy. 
% Such a collaborative approach is especially beneficial in challenging scenarios where a single model may struggle, allowing the ensemble to make more reliable and confident decisions.
% This ensemble method effectively leverages the complementary strengths of multiple large models by fusing their soft predictions, enhancing robustness and improving detection accuracy.

%% file: tables/real2.tex
\begin{table*}
\setlength{\belowcaptionskip}{-0.01cm}
\centering
\belowrulesep=0pt
\aboverulesep=0pt
\renewcommand\arraystretch{0.8}
\vspace{-2mm}
\caption{{\bf Performance benchmark on real image subsets.} $\heartsuit$Conventional deepfake detection models, $\bigstar$open-source and $\triangle$close-source LMMs. $\blacklozenge$* refers to finetuned models.  {\red{Best}} and \blue{second-best} zero-shot results. 
\colorbox{red!20}{Best} and \colorbox{mycolor_blue!200}{second-best} finetuned results.
% All experiments for AQA and VQA methods are retrained on each dimension under 10 random train-test splits at a ratio of 8:2.
}\vspace{-4mm}
   \resizebox{\linewidth}{!}{\begin{tabular}{l|c|c|c|c|c|c|c|c}
    \toprule[1pt]
     %\multicolumn{2}{c}{\bf Training/Testing}&\multicolumn{8}{c}{{\bf GAIA} (9180)}&-\\
     %\midrule
   {\bf  Methods / Datasets}  & \textbf{LIVE}~\cite{sheikh2006statistical}&\textbf{CSIQ}~\cite{larson2010most}&\textbf{TID2013}~\cite{ponomarenko2015image}&\textbf{KADID}~\cite{lin2019kadid}&\textbf{CLIVE}~\cite{ghadiyaram2015massive}&\textbf{KonIQ-10k}~\cite{hosu2020koniq}&\textbf{Flickr8k}~\cite{hodosh2013framing}&\textbf{Overall}\\
   \cdashline{2-8}
   % \textit{mix with} & \multicolumn{2}{c:}{\textit{Fully AI}} & \multicolumn{2}{c|}{\textit{Partial AI}}& \multicolumn{2}{c:}{\textit{Fully AI}} & \multicolumn{2}{c|}{\textit{Partial AI}}& \multicolumn{2}{c:}{\textit{Fully AI}} & \multicolumn{2}{c|}{\textit{Partial AI}}&\multicolumn{2}{c:}{\textit{Fully AI}} & \multicolumn{2}{c}{\textit{Partial AI}}&\multicolumn{2}{c:}{\textit{Fully AI}} & \multicolumn{2}{c}{\textit{Partial AI}}\\
   % \cdashline{2-15}
% {\bf Methods / Metrics}&Acc$\uparrow$&F1$\uparrow$&Acc$\uparrow$&F1$\uparrow$&Acc$\uparrow$&F1$\uparrow$&Acc$\uparrow$&F1$\uparrow$&Acc$\uparrow$&F1$\uparrow$&Acc$\uparrow$&F1$\uparrow$&Acc$\uparrow$&F1$\uparrow$\\
    \midrule
   $\heartsuit$CnnSpott~\cite{DBLP:journals/corr/abs-2104-02984} &\red{99.80\%}&96.89\%&\blue{99.37\%}&\red{99.83\%}&99.74\%&\blue{99.72\%}&99.65\%&\blue{99.29\%}\\
   $\heartsuit$AntifakePrompt~\cite{DBLP:journals/corr/abs-2310-17419}&81.36\%&70.44\%&93.23\%&93.02\%&68.52\%&81.82\%&89.17\%&82.51\%\\
   $\heartsuit$Gram-Net~\cite{DBLP:conf/cvpr/LiuQT20}&86.97\%&75.78\%&93.30\%&83.40\%&86.83\%&85.21\%&84.29\%&85.11\%\\
   $\heartsuit$UnivFD~\cite{DBLP:conf/cvpr/OjhaLL23}&91.24\%&86.22\%&92.43\%&91.79\%&99.66\%&98.38\%&99.83\%&94.22\%\\
   $\heartsuit$LGrad~\cite{DBLP:conf/cvpr/Tan0WGW23}&82.89\%&54.22\%&99.53\%&70.14\%&72.88\%&77.14\%&45.83\%&71.81\%\\
   \hdashline
   $\bigstar$Llava-one-vision (0.5B)~\cite{xiong2024llavaovchat} &\red{99.80\%}&\red{98.44\%}&\red{100.0\%}&\blue{98.23\%}&\red{100.0\%}&\red{99.88\%}&99.96\%&\red{99.47\%}\\
    $\bigstar$DeepSeekVL (7B)~\cite{lu2024deepseekvl}&88.90\%&87.11\%&79.43\%&83.95\%&96.82\%&97.83\%&99.68\%&90.53\%\\
    $\bigstar$LLaVA-1.5 (7B)~\cite{liu2024improved} &93.38\%&96.33\%&91.30\%&95.72\%&\red{100.0\%}&99.28\%&\red{100.0\%}&96.57\%\\
    $\bigstar$Llava-one-vision (7B)~\cite{xiong2024llavaovchat}&79.53\%&76.33\%&74.43\%&77.47\%&99.40\%&98.52\%&99.91\%&86.51\%\\
    $\bigstar$mPLUG-Owl3 (7B)~\cite{ye2024mplug}&75.76\%&76.44\%&65.41\%&65.32\%&97.59\%&96.17\%&99.73\%&82.35\% \\ 
 $\bigstar$Qwen2.5-VL (7B)~\cite{Qwen2.5-VL}&77.70\%&76.44\%&72.00\%&74.89\%&96.39\%&96.20\%&99.23\%&84.69\%\\
  
  $\bigstar$CogAgent (18B)~\cite{hong2024cogagentvisuallanguagemodel}&81.98\%&91.11\%&77.83\%&84.41\%&99.40\%&98.67\%&99.86\%&90.47\%\\
    $\bigstar$InternVL2.5 (8B)~\cite{chen2024expanding}&78.21\%&80.67\%&75.67\%&70.96\%&95.78\%&94.14\%&99.69\%&85.02\%\\
    $\bigstar$InternVL3 (9B)~\cite{wang2024mpo}&72.40\%&73.89\%&63.83\%&60.77\%&94.23\%&92.80\%&99.62\%&79.65\%\\
   $\bigstar$InternLM-XComposer2.5 (7B)~\cite{internlmxcomposer}&90.43\%&93.89\%&89.60\%&92.81\%&99.83\%&99.33\%&100.0\%&96.05\%\\
   $\bigstar$LLaVA-NeXT (8B)~\cite{liu2024llavanext}&74.85\%&73.22\%&68.13\%&75.81\%&90.71\%&84.94\%&98.16\%&80.83\%\\
    $\bigstar$Llama3.2-Vision (11B)~\cite{meta2024llama}&53.56\%&49.78\%&50.52\%&45.65\%&66.67\%&59.35\%&68.39\%&56.27\%\\
   $\bigstar$Qwen2-VL (72B) \cite{wang2024qwen2}&77.60\%&76.80\%&73.20\%&89.60\%&95.40\%&48.43\%&97.20\%&86.40\%\\
   $\bigstar$Qwen2.5-VL (72B)~\cite{Qwen2.5-VL}&79.23\%&71.80\%&71.60\%&90.60\%&96.80\%&48.35\%&98.40\%&86.43\%\\
   $\bigstar$Llava-one-vision (72B)~\cite{xiong2024llavaovchat}&77.49\%&74.20\%&70.60\%&92.75\%&99.20\%&48.32\%&99.60\%&87.35\%\\
   $\bigstar$InternVL2.5 (78B)~\cite{chen2024expanding}&67.82\%&68.60\%&64.40\%&84.00\%&94.00\%&47.19\%&95.60\%&80.17\%\\
   $\bigstar$InternVL3 (78B)~\cite{wang2024mpo}&69.25\%	&96.00\%	&64.60\%	&86.00\%	&96.00\%	&92.60\%	&97.80\%	&86.04\%
\\
    \hdashline
     $\triangle$Gemini1.5-pro \cite{Gemini}&97.45\%&\blue{97.00\%}&91.96\%&91.70\%&\red{100.0\%}&99.60\%&\red{100.0\%}&96.82\%\\
     
     $\triangle$Grok2 Vision \cite{Grok2}&76.48\%&74.67\%&63.27\%&67.22\%&76.48\%&94.31\%&98.28\%&78.67\%\\
 \hdashline
  \rowcolor{mycolor_green}\textbf{Model Average (Zero-shot)} &81.42\% &	79.85\%	&78.57\%	&81.92\% &	92.60\% &	79.14\%	 &94.58\%	&84.91\%
\\
\hline
$\blacklozenge$LGrad*~\cite{DBLP:conf/cvpr/Tan0WGW23}&56.52\%	&53.33\%	&99.00\%	&97.85\%	&68.24\%	&92.53\%&	81.94\%	&78.49\%
\\
 $\blacklozenge$InternVL2.5* (8B)~\cite{chen2024expanding}&\cellcolor{red!20}91.30\%	&\cellcolor{mycolor_blue!200}99.45\% &	\cellcolor{mycolor_blue!200}99.83\%	& \cellcolor{mycolor_blue!200}99.85\%	& 95.34\%	&\cellcolor{mycolor_blue!200}99.86\%	&99.20\%	&\cellcolor{mycolor_blue!200}97.83\%
\\
 $\blacklozenge$InternVL3* (9B)~\cite{wang2024mpo}&83.33\%&	98.91\%&	99.02\%	&99.81\%	&\cellcolor{mycolor_blue!200}96.17\%	&99.38\%&	\cellcolor{red!20}99.94\%	&96.65\%
\\
 $\blacklozenge$Qwen2.5-VL* (7B)~\cite{Qwen2.5-VL}&52.17\%	&98.33\%&	100.0\%	&99.17\%&	90.56\%&	98.84\%&	92.16\%	&90.18\%
\\
 $\blacklozenge$\textbf{MoA-DF (Ours)}&\cellcolor{mycolor_blue!200}90.91\%&	\cellcolor{red!20}100.0\%&	\cellcolor{red!20}100.0\%	&\cellcolor{red!20}99.90\%	&\cellcolor{red!20}98.30\%	&\cellcolor{red!20}99.91\%	&\cellcolor{mycolor_blue!200}99.82\%	&\cellcolor{red!20}98.41\%
 \\

    \bottomrule[1pt]
  \end{tabular}}
  \label{real}
  % \vspace{-2em}
\end{table*}

%% file: tables/partial.tex
\begin{table*}
\setlength{\belowcaptionskip}{-0.01cm}
\centering
\belowrulesep=0pt
\aboverulesep=0pt
\renewcommand\arraystretch{0.8}
\vspace{-3mm}
\caption{{\bf Performance benchmark on AI-edit subsets, including real source and four editing types.} $\blacklozenge$* refers to finetuned models. 
% All experiments for AQA and VQA methods are retrained on each dimension under 10 random train-test splits at a ratio of 8:2.
}\vspace{-4mm}
   \resizebox{\linewidth}{!}{\begin{tabular}{l|cc|cc|cc|cc|cc|cc}
    \toprule[1pt]
     %\multicolumn{2}{c}{\bf Training/Testing}&\multicolumn{8}{c}{{\bf GAIA} (9180)}&-\\
     %\midrule
   {\bf Dimension}  &\multicolumn{2}{c|}{\textbf{Object Enhance}}&\multicolumn{2}{c|}{ \textbf{Object Operation}}&\multicolumn{2}{c|}{\textbf{Semantic Change}}& \multicolumn{2}{c|}{\textbf{Style Change}}& \multicolumn{2}{c|}{\textbf{Real Source}}&\multicolumn{2}{c}{\textbf{Overall}}\\
   \cdashline{2-13}
   % \textit{mix with} & \multicolumn{2}{c:}{\textit{Fully AI}} & \multicolumn{2}{c|}{\textit{Real}}& \multicolumn{2}{c:}{\textit{Fully AI}} & \multicolumn{2}{c|}{\textit{Real}}& \multicolumn{2}{c:}{\textit{Fully AI}} & \multicolumn{2}{c|}{\textit{Real}}&\multicolumn{2}{c:}{\textit{Fully AI}} & \multicolumn{2}{c}{\textit{Real}}&\multicolumn{2}{c:}{\textit{Fully AI}} & \multicolumn{2}{c}{\textit{Real}}\\
   % \cdashline{2-15}
{\bf Methods / Metrics}&Acc(\%)$\uparrow$&F1$\uparrow$&Acc(\%)$\uparrow$&F1$\uparrow$&Acc(\%)$\uparrow$&F1$\uparrow$&Acc(\%)$\uparrow$&F1$\uparrow$&Acc(\%)$\uparrow$&F1$\uparrow$&Acc(\%)$\uparrow$&F1$\uparrow$\\
    \midrule
    $\heartsuit$CnnSpott~\cite{DBLP:journals/corr/abs-2104-02984}&0.661&0.013 &0.861&0.017 &1.131&0.022 &1.184&0.023 &\red{99.90}&0.629 &50.43&0.324 \\
     $\heartsuit$AntifakePrompt~\cite{DBLP:journals/corr/abs-2310-17419}&43.64&0.479 &40.15&0.443 &24.14&0.339 &25.36&0.340 &65.32&0.531 &49.32&0.465 \\
      $\heartsuit$Gram-Net~\cite{DBLP:conf/cvpr/LiuQT20}&10.66&0.176 &11.29&0.185 &9.641&0.164 &9.389&0.161 &89.28&0.603 &49.76&0.387 \\
      $\heartsuit$UnivFD~\cite{DBLP:conf/cvpr/OjhaLL23}&5.372&0.101 &9.103&0.166 &12.52&0.220 &14.02&0.243 &98.54&0.646 &54.39&0.414\\
      $\heartsuit$LGrad~\cite{DBLP:conf/cvpr/Tan0WGW23}&\red{52.07}&\red{0.577} &\red{59.15}&\red{0.627}&\red{63.36}&\red{0.729} &\red{63.98}&\blue{0.708} &76.06&0.679 &\blue{67.85}&\blue{0.670} \\
      \hdashline

      $\bigstar$Llava-one-vision (0.5B)~\cite{xiong2024llavaovchat}&0.820&0.016 &0.983&0.019 &2.439&0.048 &1.611&0.032 &\blue{99.80}&0.627 &50.63&0.328 \\
      
    $\bigstar$DeepSeekVL (7B)~\cite{lu2024deepseekvl}&1.983&0.037 &3.509&0.065 &6.507&0.119 &7.560&0.136 &95.85&0.621 &50.37&0.355 \\
     
    $\bigstar$LLaVA-1.5 (7B)~\cite{liu2024improved}&0.909&0.018 &1.225&0.024 &2.856&0.055 &3.484&0.067 &99.45&0.630 &50.78&0.335  \\
    $\bigstar$Llava-one-vision (7B)~\cite{xiong2024llavaovchat}&4.711&0.087 &6.620&0.121 &13.97&0.241 &16.77&0.281 &96.97&0.640 &53.74&0.411\\
    $\bigstar$mPLUG-Owl3 (7B)~\cite{ye2024mplug}&5.207&0.092 &6.819&0.120 &14.96&0.247 &17.39&0.280 &91.69&0.619 &51.39&0.402 \\ 
 $\bigstar$Qwen2.5-VL (7B)~\cite{Qwen2.5-VL}&18.10&0.291 &21.28&0.334 &35.15&0.500 &38.04&0.527 &92.16&0.651 &60.15&0.532\\
  
  $\bigstar$CogAgent (18B)~\cite{hong2024cogagentvisuallanguagemodel}&6.116&0.112 &9.864&0.176 &19.62&0.324 &21.60&0.349 &97.23&0.651 &55.77&0.446 \\
    $\bigstar$InternVL2.5 (8B)~\cite{chen2024expanding}&10.80&0.182 &12.62&0.212 &20.47&0.324 &21.34&0.335 &91.49&0.611 &53.90&0.437 \\
    $\bigstar$InternVL3 (9B)~\cite{wang2024mpo}&14.07&0.229 &15.87&0.255 &23.58&0.361 &24.93&0.376 &89.43&0.606 &54.52&0.456 \\
   $\bigstar$InternLM-XComposer2.5 (7B)~\cite{internlmxcomposer}&0.530&0.010 &0.572&0.011 &1.047&0.020 &1.499&0.029 &97.86&0.626 &49.39&0.322  \\
   $\bigstar$LLaVA-NeXT (8B)~\cite{liu2024llavanext}&34.05&0.433 &36.31&0.469 &43.53&0.546 &43.80&0.545 &77.96&0.624 &58.69&0.561 \\
    $\bigstar$Llama3.2-Vision (11B)~\cite{meta2024llama}&\blue{43.68}&0.484 &43.26&0.481 &48.56&0.551 &49.88&0.557 &61.35&0.544 &53.85&0.531 \\
   $\bigstar$Qwen2-VL (72B) \cite{wang2024qwen2}&8.683&0.143 &15.97&0.249 &24.63&0.364 &27.89&0.402 &91.20&0.724 &55.25&0.507 \\
   $\bigstar$Qwen2.5-VL (72B)~\cite{Qwen2.5-VL}&12.28&0.193 &18.99&0.286 &27.41&0.384 &32.04&0.437 &88.74&0.719 &55.71&0.522\\
   $\bigstar$Llava-one-vision (72B)~\cite{xiong2024llavaovchat}&8.982&0.150 &14.84&0.241 &26.95&0.395 &31.64&0.454 &92.99&\red{0.736} &56.80&0.523 \\
   $\bigstar$InternVL2.5 (78B)~\cite{chen2024expanding}&35.76&0.404 &40.63&0.460&55.22&0.603 &57.63&0.606&74.02&0.693 &60.66&0.606 \\
   $\bigstar$InternVL3 (78B)~\cite{wang2024mpo}&20.06&0.276&28.68&0.386&38.89&0.494&42.04&0.523&84.48&0.719&58.45&0.569\\
    \hdashline
     $\triangle$Gemini1.5-pro \cite{Gemini}&1.321&0.026 &3.257&0.063 &4.501&0.086 &6.374&0.120 &99.56&0.634 &51.71&0.354\\
     
     $\triangle$Grok2 Vision \cite{Grok2}&40.83&\blue{0.533} &\blue{46.48}&\blue{0.597} &\blue{60.19}&\blue{0.714} &\blue{62.22}&\red{0.727} &88.99&\blue{0.735} &\red{70.71}&\red{0.689} \\
     \hdashline
      \rowcolor{mycolor_green}\textbf{Model Average (Zero-shot)} &15.89&0.211&18.68&0.250&24.22&0.327&25.90&34.41&89.18&0.646&55.18&0.464\\
      \hline
      $\blacklozenge$LGrad*~\cite{DBLP:conf/cvpr/Tan0WGW23}&68.24	&0.742 &	65.54	&0.711 	&79.32&	0.829 &	75.48	&0.801 	&72.15&	0.771 &	77.39	&0.770 
\\
 $\blacklozenge$InternVL2.5* (8B)~\cite{chen2024expanding}&\cellcolor{red!20}97.27	&\cellcolor{red!20}0.976 	&\cellcolor{mycolor_blue!200}92.98	&\cellcolor{mycolor_blue!200}0.952 &	\cellcolor{mycolor_blue!200}96.82	&\cellcolor{mycolor_blue!200}0.976 	&\cellcolor{mycolor_blue!200}96.34&	\cellcolor{mycolor_blue!200}0.974 &	\cellcolor{mycolor_blue!200}95.85	&\cellcolor{mycolor_blue!200}0.970 &	\cellcolor{mycolor_blue!200}96.84	&\cellcolor{mycolor_blue!200}0.968 
\\
 $\blacklozenge$InternVL3* (9B)~\cite{wang2024mpo}&92.16	&0.934 	&85.92&	0.896& 	96.30&	0.965 &	93.88	&0.950 &	92.06	&0.936 	&93.46	&0.934 
\\
 $\blacklozenge$Qwen2.5-VL* (7B)~\cite{Qwen2.5-VL}&81.57	&0.847 &	85.12	&0.879 	&93.62	&0.931 	&93.00&	0.928 &	88.33	&0.896& 	89.08	&0.891 
\\
$\blacklozenge$\textbf{MoA-DF (Ours) }&\cellcolor{mycolor_blue!200}96.08	&\cellcolor{mycolor_blue!200}0.972 &	\cellcolor{red!20}93.27	&\cellcolor{red!20}0.955 &	\cellcolor{red!20}97.85	&\cellcolor{red!20}0.983 &	\cellcolor{red!20}96.51	&\cellcolor{red!20}0.976 &	\cellcolor{red!20}95.93	&\cellcolor{red!20}0.971 &	\cellcolor{red!20}97.07	&\cellcolor{red!20}0.970 
\\

     % $\blacklozenge$Qwen2.5-VL* (7B)~\cite{Qwen2.5-VL} &-\\
   
% $\bigstar$\textbf{LLaVA-1.5 (7B)}~\cite{wang2025internvideo}
% & 0.3372 & 0.3525 & 0.2577 & 0.3887 
% &-\\
% $\bigstar$\textbf{LLaVA-NeXT (8B)}~\cite{liu2024llavanext}
% & 0.4333 & 0.4164 & 0.3442 & 0.4568 

% &-\\
% $\bigstar$\textbf{mPLUG-Owl3 (7B)}~\cite{ye2024mplug}
% & 0.3918 & 0.3569 & 0.3018 & 0.4744 
% &-\\
% $\bigstar$\textbf{MiniCPM-V2.6 (8B)}~\cite{damonlpsg2025videollama3}
% & 0.3733 & 0.1053 & 0.2839 & 0.5916 
% &-\\
% $\bigstar$\textbf{Qwen2-VL (7B)}~\cite{Qwen2-VL}
% & 0.3760 & 0.3625 & 0.3061 & 0.5899
% &-\\
% % $\clubsuit$\textbf{ MiniCPM-V2.6} \cite{}
% % & 0.3733 & 0.1053 & 0.2839 & 0.5916 & 0.5971 & 0.4597
% % &-\\

% $\bigstar$ \textbf{DeepSeekVL (7B)}~\cite{wu2024deepseekvl2mixtureofexpertsvisionlanguagemodels}
% & 0.2611 & 0.3010 & 0.1988
% & 0.2356 
 
% &-\\
    \bottomrule[1pt]
  \end{tabular}}
  \label{partial}
  \vspace{-3mm}
\end{table*}

%% file: tables/Acc.tex
\begin{table*}
\setlength{\belowcaptionskip}{-0.01cm}
\centering
\belowrulesep=0pt
\aboverulesep=0pt
\renewcommand\arraystretch{0.8}
% \vspace{-6mm}
\caption{Performance benchmark on AI-generated subsets. $\heartsuit$Deepfake detection models, $\bigstar$open-source and $\triangle$close-source LMMs. $\blacklozenge$* refers to finetuned models.  {\red{Best}} and \blue{second-best} zero-shot results. 
\colorbox{red!20}{Best} and \colorbox{mycolor_blue!200}{second-best} finetuned results.
}
\vspace{-3mm}
   \resizebox{\linewidth}{!}{\begin{tabular}{l|cc|cc|cc|cc|cc|cc|cc}
    \toprule[1pt]
     %\multicolumn{2}{c}{\bf Training/Testing}&\multicolumn{8}{c}{{\bf GAIA} (9180)}&-\\
     %\midrule
   {\bf Datasets}  & \multicolumn{2}{c|}{\textbf{Playground}}&\multicolumn{2}{c|}{\textbf{SD3.5\_Large}}&\multicolumn{2}{c|}{ \textbf{PixArt-Sigma}}&\multicolumn{2}{c|}{\textbf{Infinity}}&\multicolumn{2}{c|}{\textbf{Kandinsky-3}}& \multicolumn{2}{c|}{\textbf{Flux\_Schnell}}& \multicolumn{2}{c}{\textbf{Kolors}}\\
   \cdashline{2-15}
   % \textit{mix with} & \multicolumn{2}{c:}{\textit{Fully AI}} & \multicolumn{2}{c|}{\textit{Real}}& \multicolumn{2}{c:}{\textit{Fully AI}} & \multicolumn{2}{c|}{\textit{Real}}& \multicolumn{2}{c:}{\textit{Fully AI}} & \multicolumn{2}{c|}{\textit{Real}}&\multicolumn{2}{c:}{\textit{Fully AI}} & \multicolumn{2}{c}{\textit{Real}}&\multicolumn{2}{c:}{\textit{Fully AI}} & \multicolumn{2}{c}{\textit{Real}}\\
   % \cdashline{2-15}
{\bf Methods / Metrics}&Acc(\%)$\uparrow$&F1$\uparrow$&Acc(\%)$\uparrow$&F1$\uparrow$&Acc(\%)$\uparrow$&F1$\uparrow$&Acc(\%)$\uparrow$&F1$\uparrow$&Acc(\%)$\uparrow$&F1$\uparrow$&Acc(\%)$\uparrow$&F1$\uparrow$&Acc(\%)$\uparrow$&F1$\uparrow$\\
    \midrule
   $\heartsuit$CnnSpott~\cite{DBLP:journals/corr/abs-2104-02984} &0.000&0.000&0.363&0.007&0.000&0.000&0.000&0.000&0.000&0.000&0.000&0.000&0.000&0.000\\
   $\heartsuit$AntifakePrompt~\cite{DBLP:journals/corr/abs-2310-17419} &0.913&0.016&4.775 &0.083 &1.113&0.020&0.638&0.011&1.625&0.029&7.763&0.131&1.150&0.021\\
   $\heartsuit$Gram-Net~\cite{DBLP:conf/cvpr/LiuQT20}&10.98&0.173&4.800&0.080&2.213&0.375&2.425&0.041&0.113&0.001&0.050&0.001&0.025&0.000\\
   $\heartsuit$UnivFD~\cite{DBLP:conf/cvpr/OjhaLL23}&0.063&0.001 &0.100&0.002  &0.563&0.011 &0.063&0.001 &0.113&0.002 &0.050&0.001 &0.025&0.000\\
   $\heartsuit$LGrad~\cite{DBLP:conf/cvpr/Tan0WGW23}&70.54&0.626 &\red{70.73}&\blue{0.627} &35.89&0.376 &\blue{89.94}&0.735 &\red{87.74}&0.724 &\blue{67.29}&0.606 &5.088&0.064\\
   \hdashline
   $\bigstar$Llava-one-vision (0.5B)~\cite{xiong2024llavaovchat} &0.000&0.000 &0.000&0.000 &0.000&0.000 &0.000&0.000 &0.000&0.000 &0.000&0.000 &0.000&0.000 \\
    % $\bigstar$DeepseekVL2 (1B)~\cite{wu2024deepseekvl2mixtureofexpertsvisionlanguagemodels}\\
    $\bigstar$DeepSeekVL (7B)~\cite{lu2024deepseekvl}&8.513&0.156 &0.725&0.014 &16.54&0.283 &3.488&0.067 &14.75&0.256 &1.525&0.030 &5.763&0.109 \\
    $\bigstar$LLaVA-1.5 (7B)~\cite{liu2024improved}&2.038&0.040 &0.113&0.002 &3.375&0.065 &1.875&0.037 &13.21&0.233 &0.438&0.009 &1.413&0.028  \\
    $\bigstar$Llava-one-vision (7B)~\cite{xiong2024llavaovchat}& 12.88&0.228 &1.063&0.021 &24.15&0.390 &6.050&0.114 &16.64&0.285 &4.488&0.086 &8.850&0.162 \\
    $\bigstar$mPLUG-Owl3 (7B)~\cite{ye2024mplug} &19.66&0.328 &2.213&0.043 &42.15&0.592 &16.94&0.289 &19.30&0.323 &13.03&0.230 &20.48&0.339\\ 
   % $\bigstar$Qwen2-VL (7B) \cite{wang2024qwen2}\\
 $\bigstar$Qwen2.5-VL (7B)~\cite{Qwen2.5-VL}&54.94 &0.706&15.46&0.266&66.60&0.796&27.44&0.428 &34.59& 0.511& 28.60 &0.442 & 45.94 & 0.626\\
   
  $\bigstar$CogAgent (18B)~\cite{hong2024cogagentvisuallanguagemodel}&10.86&0.196 &1.175&0.023 &23.80&0.384 &3.763&0.072 &16.76&0.287 &1.463&0.029 &8.100&0.150 \\
    $\bigstar$InternVL2.5 (8B)~\cite{chen2024expanding} &40.13&0.571 &5.013&0.095 &44.44&0.614 &17.78&0.301 &27.70&0.433 &13.30&0.234 &20.39&0.338\\
    $\bigstar$InternVL3 (9B)~\cite{wang2024mpo}&41.28&0.583 &6.325&0.119 &48.66&0.653 &22.63&0.368 &30.24&0.463 &13.69&0.240 &26.88&0.422 \\
   $\bigstar$InternLM-XComposer2.5 (7B)~\cite{internlmxcomposer}&15.69&0.294 &1.150&0.023 &19.51&0.327 &4.025&0.077 &16.20&0.279 &1.875&0.037 &8.438&0.156 \\
   $\bigstar$LLaVA-NeXT (8B)~\cite{liu2024llavanext}&30.16&0.457 &4.738&0.089 &35.21&0.514 &10.00&0.179 &23.79&0.379 &4.550&0.086 &18.78&0.311 \\
   % $\bigstar$InternVL2.5 (38B)~\cite{chen2024expanding}\\
   % $\bigstar$InternVL3 (38B)~\cite{wang2024mpo}\\
   $\bigstar$Llama3.2-Vision (11B)~\cite{meta2024llama}&\red{90.21}&\blue{0.816} &\blue{64.08}&\red{0.658} &\blue{89.08}&0.812 &\red{91.29}&0.818 &\blue{81.50}&0.765 &\red{81.95}&\red{0.768} &\blue{83.64}&0.796 \\
   $\bigstar$Qwen2-VL (72B) \cite{wang2024qwen2}&45.60&0.624 &14.00&0.244 &53.20&0.692 &42.80&0.597 &43.80&0.607 &27.20&0.426 &40.00&0.569\\
   $\bigstar$Qwen2.5-VL (72B)~\cite{Qwen2.5-VL}&\blue{86.60}&\red{0.920} &19.60&0.323 &\red{92.40}&\red{0.953} &72.00&\blue{0.829} &64.60&\blue{0.777} &52.00&\blue{0.677} &\red{91.00}&\red{0.945}\\
   $\bigstar$Llava-one-vision (72B)~\cite{xiong2024llavaovchat}&33.40&0.499 &4.200&0.080 &48.60&0.652 &25.80&0.409 &26.80&0.421 &18.80&0.315 &22.00&0.359\\
   $\bigstar$InternVL2.5 (78B)~\cite{chen2024expanding}&69.60&0.800 &29.80&0.444 &87.40&\blue{0.911} &81.00&\red{0.874} &74.80&\red{0.835} &52.20&0.667 &78.80&\blue{0.860} \\
   $\bigstar$InternVL3 (78B)~\cite{wang2024mpo}&41.28 & 0.583 & 6.235 & 0.119 & 48.66 & 0.653 & 22.63& 0.368 &30.24 & 0.463 & 13.69 &0.240 &26.88 &0.422\\
    \hdashline
     $\triangle$Gemini1.5-pro \cite{Gemini}&9.538&0.175 &0.675&0.013 &17.39&0.297 &4.663&0.089 &14.83&0.258 &1.250&0.025 &6.800&0.128 \\
     
     $\triangle$Grok2 Vision \cite{Grok2}&32.45&0.484 &11.96&0.211 &46.23&0.625 &23.44&0.375 &30.86&0.466 &19.96&0.328 &58.64&0.451\\

\hdashline
\rowcolor{mycolor_green}\textbf{Model Average (Zero-shot)} &30.30 &0.387 &11.22& 0.149 &35.30 &0.444& 23.78&0.295 &27.92 &0.367 &18.10 & 0.240 &24.22 & 0.304\\
\hline
$\blacklozenge$LGrad*~\cite{DBLP:conf/cvpr/Tan0WGW23}&98.69&	0.904 
  &98.11&	0.901 &	95.75&	0.889 &	99.88&	0.910 &	100.0&	0.911& 	98.25&	0.902& 	98.19&	0.902 
\\
 $\blacklozenge$InternVL2.5* (8B)~\cite{chen2024expanding}& \cellcolor{red!20}\cellcolor{red!20}100.0&	0.996 &	99.81&	0.995& 	\cellcolor{red!20}100.0&	0.996 &	\cellcolor{red!20}100.0&	0.996 &	100.0&	0.996& 	99.94&	0.996 &	\cellcolor{red!20}100.0&	0.996 
\\
 $\blacklozenge$InternVL3* (9B)~\cite{wang2024mpo}&99.88&	 \cellcolor{red!20}0.999 	&99.25&	\cellcolor{mycolor_blue!200}0.996 &	99.81&	\cellcolor{red!20}0.999 &	99.88&	\cellcolor{red!20}0.999& 	99.94&	\cellcolor{red!20}0.999 &	99.44&	\cellcolor{mycolor_blue!200}0.997& 	\cellcolor{red!20}100.0&	\cellcolor{red!20}1.000 
\\
 $\blacklozenge$Qwen2.5-VL* (7B)~\cite{Qwen2.5-VL} & \cellcolor{red!20}100.0&	0.962 &	\cellcolor{red!20}99.94&	0.961 &	\cellcolor{red!20}100.0&	0.962 &	\cellcolor{red!20}100.0&	0.962 	&\cellcolor{red!20}100.0&	0.962 &	\cellcolor{red!20}100.0&	0.962 &	\cellcolor{red!20}100.0&	0.962 
\\
 $\blacklozenge$\textbf{MoA-DF (Ours)}& \cellcolor{red!20}100.0&	\cellcolor{mycolor_blue!200}0.998 &	\cellcolor{mycolor_blue!200}99.85&	\cellcolor{red!20}0.997 	&\cellcolor{red!20}100.0&	\cellcolor{mycolor_blue!200}0.998 &	\cellcolor{red!20}100.0&	\cellcolor{mycolor_blue!200}0.998 &	\cellcolor{red!20}100.0&	\cellcolor{mycolor_blue!200}0.998 &	\cellcolor{red!20}100.0&	\cellcolor{red!20}0.998& 	\cellcolor{red!20}100.0&	\cellcolor{mycolor_blue!200}0.998 
\\

     % $\blacklozenge$Qwen2.5-VL* (7B)~\cite{Qwen2.5-VL} &-\\
   
% $\bigstar$\textbf{LLaVA-1.5 (7B)}~\cite{wang2025internvideo}
% & 0.3372 & 0.3525 & 0.2577 & 0.3887 
% &-\\
% $\bigstar$\textbf{LLaVA-NeXT (8B)}~\cite{liu2024llavanext}
% & 0.4333 & 0.4164 & 0.3442 & 0.4568 

% &-\\
% $\bigstar$\textbf{mPLUG-Owl3 (7B)}~\cite{ye2024mplug}
% & 0.3918 & 0.3569 & 0.3018 & 0.4744 
% &-\\
% $\bigstar$\textbf{MiniCPM-V2.6 (8B)}~\cite{damonlpsg2025videollama3}
% & 0.3733 & 0.1053 & 0.2839 & 0.5916 
% &-\\
% $\bigstar$\textbf{Qwen2-VL (7B)}~\cite{Qwen2-VL}
% & 0.3760 & 0.3625 & 0.3061 & 0.5899
% &-\\
% % $\clubsuit$\textbf{ MiniCPM-V2.6} \cite{}
% % & 0.3733 & 0.1053 & 0.2839 & 0.5916 & 0.5971 & 0.4597
% % &-\\

% $\bigstar$ \textbf{DeepSeekVL (7B)}~\cite{wu2024deepseekvl2mixtureofexpertsvisionlanguagemodels}
% & 0.2611 & 0.3010 & 0.1988
% & 0.2356 
 
% &-\\
    \bottomrule[1pt]
  \end{tabular}}
  \label{mos}

     \resizebox{\linewidth}{!}{\begin{tabular}{l|cc|cc|cc|cc|cc|cc|cc}
    \toprule[1pt]
     %\multicolumn{2}{c}{\bf Training/Testing}&\multicolumn{8}{c}{{\bf GAIA} (9180)}&-\\
     %\midrule
   {\bf Datasets}  & \multicolumn{2}{c|}{\textbf{SD3\_Medium}}&\multicolumn{2}{c|}{\textbf{Flux\_dev}}&\multicolumn{2}{c|}{ \textbf{NOVA}}&\multicolumn{2}{c|}{\textbf{LaVi-Bridge}}&\multicolumn{2}{c|}{\textbf{Janus}}& \multicolumn{2}{c|}{\textbf{Real  Source}}& \multicolumn{2}{c}{\textbf{Overall}}\\
   \cdashline{2-15}
   % \textit{mix with} & \multicolumn{2}{c:}{\textit{Fully AI}} & \multicolumn{2}{c|}{\textit{Real}}& \multicolumn{2}{c:}{\textit{Fully AI}} & \multicolumn{2}{c|}{\textit{Real}}& \multicolumn{2}{c:}{\textit{Fully AI}} & \multicolumn{2}{c|}{\textit{Real}}&\multicolumn{2}{c:}{\textit{Fully AI}} & \multicolumn{2}{c}{\textit{Real}}&\multicolumn{2}{c:}{\textit{Fully AI}} & \multicolumn{2}{c}{\textit{Real}}\\
   % \cdashline{2-15}
{\bf Methods / Metrics}&Acc(\%)$\uparrow$&F1$\uparrow$&Acc(\%)$\uparrow$&F1$\uparrow$&Acc(\%)$\uparrow$&F1$\uparrow$&Acc(\%)$\uparrow$&F1$\uparrow$&Acc(\%)$\uparrow$&F1$\uparrow$&Acc(\%)$\uparrow$&F1$\uparrow$&Acc(\%)$\uparrow$&F1$\uparrow$\\
    \midrule
   $\heartsuit$CnnSpott~\cite{DBLP:journals/corr/abs-2104-02984} &0.025&0.000 &0.000&0.000 &0.275&0.005 &0.013&0.000 &0.925&0.018 &99.65&0.668 &7.789&0.054 \\
   $\heartsuit$AntifakePrompt~\cite{DBLP:journals/corr/abs-2310-17419} &4.588&0.079 &5.713&0.098 &0.963&0.017 &0.013&0.000 &0.863&0.015 &89.17&0.625 &9.176&0.088\\
   $\heartsuit$Gram-Net~\cite{DBLP:conf/cvpr/LiuQT20}&2.688&0.045 &0.700&0.012 &0.938&0.016 &0.450&0.008 &7.888&0.127 &84.29&0.603 &9.918&0.102 \\
   $\heartsuit$UnivFD~\cite{DBLP:conf/cvpr/OjhaLL23}&0.088&0.002 &0.000 &0.000 &2.638&0.051 &0.313&0.006 &29.63&0.456 &99.83&0.675 &10.27&0.093 \\
   $\heartsuit$LGrad~\cite{DBLP:conf/cvpr/Tan0WGW23}&\blue{47.16}&0.467 &\blue{85.34}&0.711 &23.60&0.265 &81.71&0.693 &25.18&0.280 &45.83&0.488 &56.62&0.512\\
   \hdashline
   $\bigstar$Llava-one-vision (0.5B)~\cite{xiong2024llavaovchat} &0.000&0.000 &0.000&0.000 &0.000&0.000 &0.000&0.000 &0.000&0.000 &0.000&0.000 &0.000&0.000 \\
    % $\bigstar$DeepseekVL2 (1B)~\cite{wu2024deepseekvl2mixtureofexpertsvisionlanguagemodels}\\
    $\bigstar$DeepSeekVL (7B)~\cite{lu2024deepseekvl}&0.688&0.014 &6.800&0.127 &10.20&0.185 &52.78&0.689 &12.35&0.219 &99.68&0.693 &17.98&0.219 \\
    $\bigstar$LLaVA-1.5 (7B)~\cite{liu2024improved}&0.125&0.002 &4.025&0.077 &4.050&0.078 &42.88&0.600 &5.450&0.103 &\red{100.0}&0.684 &13.77&0.151   \\
    $\bigstar$Llava-one-vision (7B)~\cite{xiong2024llavaovchat}& 0.000&0.000 &0.000&0.000 &0.000&0.000 &0.000&0.000 &0.225&0.004 &99.96&0.669 &7.707&0.052 \\
    $\bigstar$mPLUG-Owl3 (7B)~\cite{ye2024mplug} &2.763&0.054 &14.09&0.246 &31.16&0.474 &84.59&0.915 &36.76&0.537 &99.73&0.729 &30.99&0.392 \\ 
   % $\bigstar$Qwen2-VL (7B) \cite{wang2024qwen2}\\
 $\bigstar$Qwen2.5-VL (7B)~\cite{Qwen2.5-VL}&19.69&0.327&33.08&0.494&32.35&0.486 &86.03&0.921&26.69&0.419 & 99.23 & 0.766 & 43.89 & 0.553\\
    $\bigstar$CogAgent (18B)~\cite{hong2024cogagentvisuallanguagemodel}&1.100&0.022 &7.163&0.134 &26.84&0.423 &69.61&0.820 &47.55&0.644 &99.86&0.711 &24.47&0.300 \\
    $\bigstar$InternVL2.5 (8B)~\cite{chen2024expanding} &9.625&0.175 &18.83&0.316 &18.94&0.318 &76.71&0.867 &17.98&0.304 &99.69&0.730 &31.58&0.407 \\
    $\bigstar$InternVL3 (9B)~\cite{wang2024mpo}&9.288&0.169 &17.85&0.304 &31.00&0.472 &84.34&0.913 &23.70&0.382 &99.62&0.740 &35.04&0.448 \\
   $\bigstar$InternLM-XComposer2.5 (7B)~\cite{internlmxcomposer}&0.925&0.018 &7.213&0.135 &7.813&0.145 &31.83&0.483 &2.830&0.055 &\red{100.0}&0.712 &17.47&0.216 \\
   $\bigstar$LLaVA-NeXT (8B)~\cite{liu2024llavanext}&5.238&0.098 &7.725&0.173 &31.95&0.478 &87.09&0.922 &47.46&0.636 &98.16&0.727 &31.14&0.388  \\
   % $\bigstar$InternVL2.5 (38B)~\cite{chen2024expanding}\\
   % $\bigstar$InternVL3 (38B)~\cite{wang2024mpo}\\
    $\bigstar$Llama3.2-Vision (11B)~\cite{meta2024llama}&\red{71.14}&\red{0.702} &\red{90.80}&\blue{0.816} &\red{66.60}&0.671 &85.60&0.787 &51.20&0.560 &68.39&0.726 &\red{78.11}&\blue{0.746}  \\
   $\bigstar$Qwen2-VL (72B) \cite{wang2024qwen2}&14.80&0.256 &40.80&0.577 &37.20&0.540 &85.20&0.917 &37.40&0.542 &98.96&0.654 &44.69&0.557 \\
   $\bigstar$Qwen2.5-VL (72B)~\cite{Qwen2.5-VL}&30.00&0.456 &76.60&\red{0.860} &71.80&\red{0.828} &\blue{98.00}&\red{0.982} &25.60&0.403 &98.40&0.843 &67.58&0.754 \\
   $\bigstar$Llava-one-vision (72B)~\cite{xiong2024llavaovchat}&6.000&0.113 &13.60&0.239 &37.60&0.545 &89.00&0.940 &49.20&0.658 &99.60&0.742 &36.51&0.459 \\
   $\bigstar$InternVL2.5 (78B)~\cite{chen2024expanding}&38.40&\blue{0.538} &61.20&0.739 &\blue{63.00}&\blue{0.753} &\red{98.40}&\blue{0.970} &\red{65.40}&\red{0.770} &95.60&0.835 &\blue{68.89}&\red{0.769} \\
   $\bigstar$InternVL3 (78B)~\cite{wang2024mpo} &25.00 & 0.393 & 39.00 & 0.552 & 51.80 & 0.673 & 93.80 & 0.957 & 57.40 & 0.719 & 97.80 & 0.860 & 55.98 & 0.666\\
    \hdashline
     $\triangle$Gemini1.5-pro \cite{Gemini}&0.600&0.012 &7.475&0.139 &10.19&0.185 &45.38&0.624 &50.19&0.669 &\red{100.0}&0.702 &20.69&0.255 \\
     
     $\triangle$Grok2 Vision \cite{Grok2}&12.65&0.221 &46.38&0.374 &34.25&0.504 &74.11&0.840 &\blue{60.06}&\blue{0.742} &98.28&0.709 &42.25&0.487 \\
     \hdashline
 \rowcolor{mycolor_green}\textbf{Model Average (Zero-shot)} &12.64 & 0.174 & 24.68 & 0.303 & 25.59 & 0.351 &59.56 & 0.651 & 29.79  & 0.407 & 94.65 & 0.708 & 32.72 & 0.373
 \\    
 \hline
  $\blacklozenge$LGrad*~\cite{DBLP:conf/cvpr/Tan0WGW23}&96.31&	0.892 &	99.69&	0.909 &	91.38&	0.866 &	99.69&	0.909 &	95.63&	0.889& 	81.94&	0.890 	&96.42&	0.898 
\\
 $\blacklozenge$InternVL2.5* (8B)~\cite{chen2024expanding}&99.94&	0.996 &	\cellcolor{red!20}100.0&	0.996& 	99.81&	0.995& 	\cellcolor{red!20}100.0&	0.996 	&98.76&	0.990 &	99.20&	0.995 &	\cellcolor{mycolor_blue!200}99.80&	0.995 
\\
 $\blacklozenge$InternVL3* (9B)~\cite{wang2024mpo}&99.88&	\cellcolor{red!20}0.999& 	99.88&	\cellcolor{red!20}0.999 &	\cellcolor{mycolor_blue!200}99.83&	\cellcolor{red!20}0.999&	\cellcolor{red!20}100.0&	\cellcolor{red!20}1.000 &	99.31&	\cellcolor{mycolor_blue!200}0.996 &	\cellcolor{red!20}99.94&	\cellcolor{red!20}0.999 &	99.77&	\cellcolor{red!20}0.998 
\\
 $\blacklozenge$Qwen2.5-VL* (7B)~\cite{Qwen2.5-VL}&\cellcolor{red!20}100.0&	0.962 &	\cellcolor{red!20}100.0&	0.962 &	99.81&	0.961 	&\cellcolor{red!20}100.0&	0.962 &	\cellcolor{red!20}100.0&	0.962 &	92.16&	0.959 &	99.38&	0.962 
\\
$\blacklozenge$\textbf{MoA-DF (Ours)}&\cellcolor{red!20}100.0	&\cellcolor{mycolor_blue!200}0.998 	&\cellcolor{red!20}100.0&	\cellcolor{mycolor_blue!200}0.998 &	\cellcolor{red!20}99.85&	\cellcolor{mycolor_blue!200}0.998& 	\cellcolor{red!20}100.0&	\cellcolor{mycolor_blue!200}0.998 &	\cellcolor{mycolor_blue!200}99.69&	\cellcolor{red!20}0.997 &	\cellcolor{mycolor_blue!200}99.70&	\cellcolor{mycolor_blue!200}0.998 &	\cellcolor{red!20}99.92&	\cellcolor{red!20}0.998 
 \\
% \rowcolor{mycolor_green}\textbf{Model Average} &27.07&	0.307 &	37.22&	0.415 &	37.68&	0.453 &	66.29&	0.704& 	41.27&	0.500& 	94.64&	0.752 &	43.78&	0.473 
%  \\
     % $\blacklozenge$Qwen2.5-VL* (7B)~\cite{Qwen2.5-VL} &-\\
   
% $\bigstar$\textbf{LLaVA-1.5 (7B)}~\cite{wang2025internvideo}
% & 0.3372 & 0.3525 & 0.2577 & 0.3887 
% &-\\
% $\bigstar$\textbf{LLaVA-NeXT (8B)}~\cite{liu2024llavanext}
% & 0.4333 & 0.4164 & 0.3442 & 0.4568 

% &-\\
% $\bigstar$\textbf{mPLUG-Owl3 (7B)}~\cite{ye2024mplug}
% & 0.3918 & 0.3569 & 0.3018 & 0.4744 
% &-\\
% $\bigstar$\textbf{MiniCPM-V2.6 (8B)}~\cite{damonlpsg2025videollama3}
% & 0.3733 & 0.1053 & 0.2839 & 0.5916 
% &-\\
% $\bigstar$\textbf{Qwen2-VL (7B)}~\cite{Qwen2-VL}
% & 0.3760 & 0.3625 & 0.3061 & 0.5899
% &-\\
% % $\clubsuit$\textbf{ MiniCPM-V2.6} \cite{}
% % & 0.3733 & 0.1053 & 0.2839 & 0.5916 & 0.5971 & 0.4597
% % &-\\

% $\bigstar$ \textbf{DeepSeekVL (7B)}~\cite{wu2024deepseekvl2mixtureofexpertsvisionlanguagemodels}
% & 0.2611 & 0.3010 & 0.1988
% & 0.2356 
 
% &-\\
    \bottomrule[1pt]
  \end{tabular}}
  \label{fake}
  % \vspace{-2em}
\end{table*}

%% file: section/4_Bench.tex
% \input{figures/model}
\vspace{-3mm}
\input{figures/per1}
\section{Benchmark and Evaluation}
We benchmark and evaluate the performance of various deepfake detection models across three subsets of DFBench: real, AI-edited, and AI-generated images. The evaluation on AI-generated images also benchmarks the evasion ability of generative models, serving as a bidirectional benchmark for both generation and detection.

\vspace{-2mm}
\subsection{Experiment Setup}

We evaluate the models’ ability to correctly classify real and fake images using two standard metrics: accuracy (Acc) and F1-score. 
Accuracy is defined as the proportion of correctly identified real or fake images out of all relevant samples in the dataset, computed as:
\begin{equation}
\text{Acc} = \frac{\text{TP}}{\text{TP} + \text{FN}}
\end{equation}
TP (True Positives) denotes the number of real or fake images correctly identified by the model, while FN (False Negatives) represents the number of images incorrectly classified as the opposite category.
To provide a balanced evaluation that considers both precision and recall, we also calculate the F1-score, the harmonic mean of precision and recall, defined as:
\begin{equation}
\text{F1} = \frac{2 \times \text{Precision} \times \text{Recall}}{\text{Precision} + \text{Recall}}
\end{equation}
where precision and recall are:
\begin{equation}
\text{Precision} = \frac{\text{TP}}{\text{TP} + \text{FP}}, \quad \text{Recall} = \frac{\text{TP}}{\text{TP} + \text{FN}}
\end{equation}
For conventional deepfake detection models, we directly utilize publicly available pre-trained weights to conduct inference on the test datasets. For large multimodal models, inference is performed via a prompt-based question-answering approach. 
We fine-tune three of the LMMs with LoRA \cite{hulora} (r=8) and LGrad~\cite{DBLP:conf/cvpr/Tan0WGW23}
using the same training and testing split (4:1). We set the
number of finetuning epoch to 1 for LMMs and 50 for LGrad~\cite{DBLP:conf/cvpr/Tan0WGW23}.
The models are implemented with PyTorch and trained on a 40GB NVIDIA RTX A6000 GPU with batch size of 4. The initial learning rate is set to 1e-5 and decreased using the cosine annealing strategy.
% Specifically, the model is asked: 
% % “Is this a real image or a generated image? Just answer with A or B. A: real or B: generated.” In the case of AI-edited datasets, where images may be partially manipulated, the prompt is adjusted to: 
% “Is this a real image, a partially AI-edited image, or a fully AI-generated image? Just answer with A or B.  A: real or B: AI.”
% Importantly, our evaluation focuses exclusively on zero-shot performance (\textit{i.e.}, models are not fine-tuned or retrained on the benchmark datasets) to better assess the inherent generalization and scaling-up capabilities of each model. This approach reflects real-world scenarios where new models can be evaluated directly on existing datasets without costly retraining, enabling a more efficient and scalable assessment of emerging detection systems.

\vspace{-2mm}
\subsection{Benchmark on Real Datasets}

% The performance of various methods on real image datasets is evaluated using accuracy, which measures the proportion of correctly predicted real images out of all real samples in the dataset. Specifically, accuracy is computed as:

% This metric focuses exclusively on the model’s ability to accurately identify real images from the dataset.

From the performance results presented in Table~\ref{real}, it is evident that most models exhibit strong zero-shot identification capabilities on real image datasets. 
However, detection accuracy generally declines on datasets containing various distortions, such as CSIQ~\cite{larson2010most} and TID2013~\cite{ponomarenko2015image}, when compared to the distortion-free Flickr8k dataset~\cite{hodosh2013framing}, indicating that image degradations such as noise, blur or compression can impact model reliability and increase the chance of misclassification. 
CnnSpott~\cite{DBLP:journals/corr/abs-2104-02984} and Llava-one-vision (0.5B)~\cite{xiong2024llavaovchat} perform well on real images mainly because they tend to classify most inputs as real, but may reduce robustness in fake detection.
\vspace{-2mm}
\subsection{Benchmark on AI-edit Datasets}
We further evaluate the performance of different detection models on AI-edit subsets.
% , focusing on their ability to distinguish manipulated images from authentic ones.
As shown in Table \ref{partial}, the high performance of 
 CnnSpott~\cite{DBLP:journals/corr/abs-2104-02984} and Llava-one-vision (0.5B)~\cite{xiong2024llavaovchat}  on real source images significantly drops on AI-edited images, resulting in relatively lower F1 scores.
% This discrepancy highlights the importance of the F1 metric, which balances precision and recall, offering a more comprehensive evaluation of detection effectiveness.
The AI-edit datasets consist of four categories: object enhancement, object operation, style change, and semantic change, each posing different challenges for detection models. 
Among these, models achieve the highest average accuracy on style change and the lowest on the object enhancement category, which involves subtle modifications to the appearance of individual objects.  These results suggest that subtle changes at the object level are more difficult for detection models to identify compared to style changes. Models show significant improvements after fine-tuning, especially LMMs trained for 1 epoch outperform conventional best network LGrad~\cite{DBLP:conf/cvpr/Tan0WGW23} trained for 50 epochs, highlighting the effectiveness of LMMs in deepfake detection tasks. 

\vspace{-3mm}
\subsection{Benchmark on AI-generation Datasets}

From Table \ref{fake}, we can observe that the detection accuracy on AI-generated datasets is generally lower compared to real image datasets, highlighting the remarkable realism achieved by current generative models and their strong capability to evade detection.  Traditional deep learning-based detection models trained on specific deepfake datasets show limited zero-shot generalization, reflecting their insufficient scaling-up capacity to handle more advanced fakes.
In contrast, large multimodal models, despite lacking task-specific training for real-fake discrimination, demonstrate relatively robust zero-shot detection performance. Among these, InternVL2.5 (78B)~\cite{chen2024expanding} achieves the best results, suggesting that larger parameter scales contribute to better generalization capabilities.
On the generation side, detection accuracy also serves as an indirect measure of generative models’ evasion effectiveness. As shown in Figure~\ref{perform22}, SD3.5-Large~\cite{esser2024scaling} attains the lowest detection accuracy, indicating its superior capacity for generating highly realistic images that effectively fool detectors, while LaVi-Bridge~\cite{zhao2024bridging} exhibits the poorest evasion performance.

%% file: figures/per1.tex
\begin{figure*}[t]
    \centering
    \includegraphics[width=\linewidth]{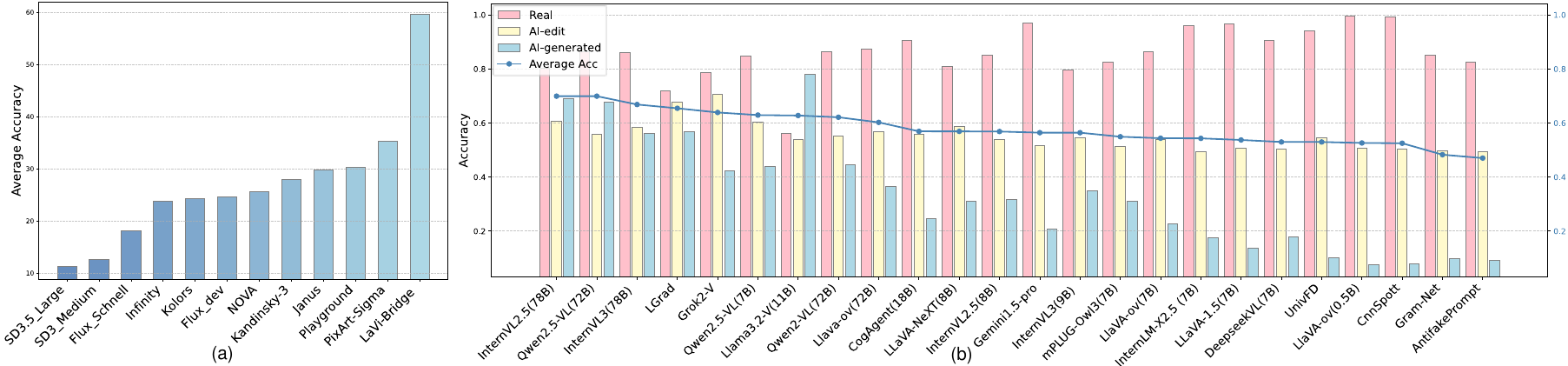}
     \vspace{-8mm}
    \caption{(a) Performance comparison of image generation models (b) Performance comparison of image detection models} 
     \vspace{-2mm}
    \label{perform22}
\end{figure*}

%% file: section/6_conclusion.tex
\vspace{-1mm}
\section{CONCLUSION}
In this paper, we introduce DFBench, a comprehensive benchmark designed to advance  deepfake image detection. DFBench features the largest scale of fake images generated by 12 state-of-the-art generative models, and rich content spanning AI-edited images and real-world image distortions.  we introduce a bidirectional evaluation protocol that assesses both the detection performance of deepfake models and the evasion strength of generative models. 
Additionally, we propose MoA-DF, a novel mixture of agents method that integrates LMMs within a unified probabilistic framework, achieving state-of-the-art performance and demonstrating the effectiveness of LMMs for deepfake detection. 
Through extensive experiments, we demonstrate the increasing realism of generative models and the limited generality of current detection methods. LMMs manifest strong zero-shot generalization ability, highlighting their potential as a promising foundation for developing more robust and generalizable deepfake detection systems.
% aims to support the development of more robust and generalizable detection systems.
% \vspace{-3mm}
% \section{Limitations and Broader Impact}
% While DFBench offers a comprehensive benchmark, several limitations remain. Some closed-source generation models with highly realistic outputs are not included due to limited access and scalability.  Several recent detection models are excluded due to unavailable weights or incomplete code. In some cases, humans can easily identify fake images that detection models misclassify as real, highlighting a perceptual gap that warrants future investigation.  Our findings suggest that detection models still have room for improvement, while generative models are advancing in evading detection. Future work could aim to enhance detection robustness and further explore how generative models can improve realism to deceive detectors more effectively.